# AOP-Smart-Finder: A High-Throughput Human–AI Hybrid Framework for AOP Discovery and Evidence Mapping from Large-Scale Literature

Qinjiang Niu[a], Lu Yan[a,*]

[a]College of Water Resources and Modern Agriculture, Nanyang Normal University, Nanyang, Henan, 473061, China

## Abstract

The rapid growth of scientific literature and the high heterogeneity of biological terminology and mechanistic descriptions make it difficult for researchers to integrate scattered research evidence, identify connections among biological events, and construct mechanistic networks. Here, we developed a high-throughput human-AI hybrid framework, AOP-Smart-Finder, for biological event discovery, relationship mining, and Adverse Outcome Pathway (AOP) construction and evidence mapping from large-scale literature. This framework integrates Large Language Models (LLMs), AOP-Smart retrieval-augmented generation (RAG) technology, and Python-based automation, supporting two analysis modes: one based on a single compound and the other based on a predefined set of biological events. In the di(2-ethylhexyl) phthalate (DEHP) case, the system processed 12,983 articles within 30 min and included 1,093 of them, from which it extracted 1,589 biological events (including 458 official key events [KEs] and 1,131 candidate KEs) and 2,519 key event relationships (KERs). In the amyotrophic lateral sclerosis (ALS) case, the system identified 55 stressors from 437 articles and extracted 278 biological events and 548 event relationships. Following manual curation and Weight of Evidence (WoE) evaluation, a mechanistic network comprising 16 KEs and 23 KERs was ultimately constructed. This framework can either use existing AOPs as knowledge guidance for mechanistic expansion and evidence supplementation, or discover novel event connections and construct mechanistic networks around user-specified research topics. It provides a high-throughput, traceable, human-AI hybrid methodology for complex biological mechanism elucidation and AOP discovery driven by large-scale literature.

## 1 Introduction

With the rapid growth of scientific literature, knowledge of biological mechanisms is increasingly dispersed across different research fields and publications.[1] Different studies often describe the same biological events with variations in terminology and expression, making it difficult for researchers to effectively integrate cross-literature evidence, identify connections among events, and construct systematic mechanistic networks.[2] Adverse Outcome Pathways (AOPs) organize biological processes through Molecular Initiating Events (MIEs), Key Events (KEs), and Key Event Relationships (KERs), providing an important framework for mechanistic knowledge integration and chemical risk assessment.[3] As an open AOP knowledge resource, AOP-Wiki continuously collects and updates relevant knowledge. However, the development and refinement of AOPs still rely heavily on manual literature screening, evidence integration, terminology standardization, and causal relationship evaluation. This process is highly labor-intensive, which not only increases the time cost of AOP development and updating but also, to some extent, limits the continuous expansion and refinement of AOP knowledge.[4]

In recent years, artificial intelligence (AI) technologies have offered new possibilities for large-scale scientific literature analysis. In the field of AOP-related literature mining, AOP-helpFinder is a representative tool. By combining natural language processing and graph theory methods, it automatically screens PubMed abstracts to identify associations between stressors and biological events, as well as among different biological events, and evaluates the strength of these associations using statistical methods.[5] However, its identification of literature associations still relies primarily on predefined term matching and co-occurrence statistics, making it difficult to fully reveal the direction of changes, the sequence of occurrence, and the specific causal mechanisms of biological events within a single publication. Therefore, how to deeply extract biological events with clear causal links from individual publications and further integrate cross-literature evidence to construct a complete mechanistic chain remains to be further explored. The development of Large Language Models (LLMs) provides a new technical pathway for this problem. Leveraging their contextual understanding and information integration capabilities, LLMs offer new approaches for identifying biological events and extracting potential relationships among events from scientific literature, and have demonstrated potential for assisting in AOP construction.[6] However, LLMs may still generate event relationships lacking textual evidence or factual errors in specialized domain knowledge extraction. Meanwhile, the dispersion of information in literature, the diversity of professional terminology, and the mapping requirements for standardized knowledge entries can also affect the consistency and accuracy of the extraction results.[7] Therefore, integrating the contextual understanding capabilities of LLMs with the retrieval of official AOP knowledge resources, and constraining the extraction results through original text evidence tracing, manual curation, and evidence evaluation, can help improve the accuracy, traceability, and reliability of large-scale mechanistic discovery.

Based on this rationale, we previously developed AOP-Smart, a RAG framework that integrates LLMs with the official AOP knowledge base.[8] By providing LLMs with accurate and standardized AOP domain knowledge, it effectively reduces the risk of hallucinations during knowledge generation. Building on this, we further developed AOP-Smart-Finder. Using AOP-Smart as the core engine connecting LLMs with official AOP-Wiki knowledge, it integrates literature screening, KER extraction, official KE mapping, semantic standardization of candidate events, and structured data output, thereby enabling high-throughput analysis from large-scale literature screening to biological mechanistic network construction. This tool aims to provide automated analytical support for cross-literature biological mechanism discovery and AOP knowledge expansion.

## 2 Materials and Methods

AOP-Smart-Finder employs a human-AI hybrid workflow, dividing tasks into manual operations, LLM tasks, joint processing by LLM and AOP-Smart, and deterministic processing by Python. The overall pipeline consists of three stages: manual literature retrieval and task definition, automated literature analysis and data output, and manual curation and WoE evaluation. Specifically, human researchers are responsible for defining research questions, formulating literature retrieval strategies, and reviewing the final results. The LLM handles literature screening, biological event identification, and KER extraction. AOP-Smart provides official AOP-Wiki knowledge support to the LLM, while Python is utilized for literature parsing, data organization, event replacement, relationship statistics, and file generation. AOP-Smart-Finder has been open-sourced on GitHub, and the source code is available at [https://github.com/qinjiang-lab/AOP-Smart-Finder](https://github.com/qinjiang-lab/AOP-Smart-Finder).

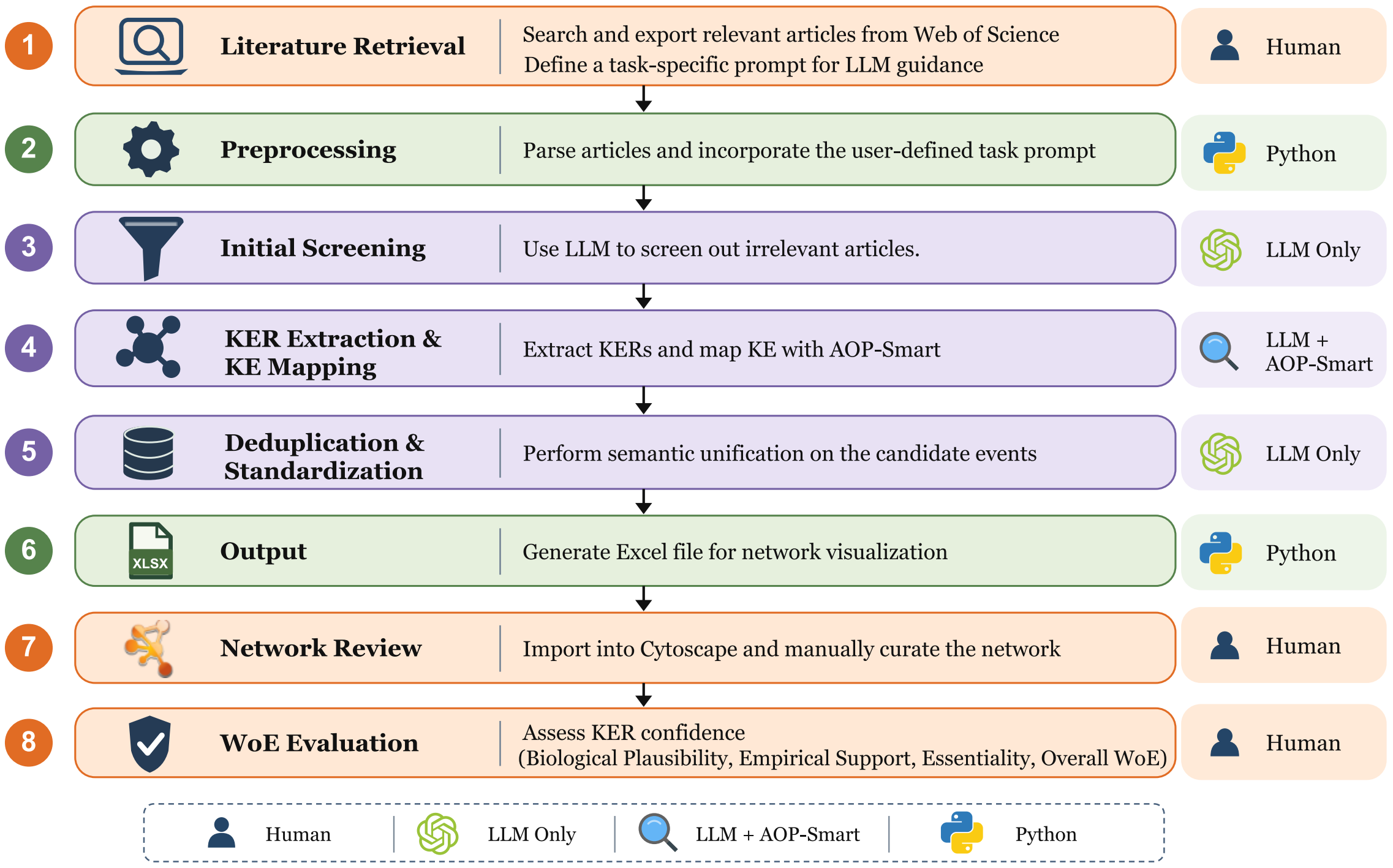

**Figure 1.** Overall workflow and task allocation of AOP-Smart-Finder

## 2.1 Literature Retrieval and Task Definition

Literature data were retrieved from the Web of Science. Researchers formulated literature retrieval strategies based on specific research objectives and exported the literature records in Excel format. The exported records must contain at least three fields: Title, Source, and Abstract, while other information can be exported as needed. The system's subsequent primary text analysis relies on the titles and abstracts.

AOP-Smart-Finder offers two analysis modes, each tailored to different research objectives.

**Mode 1**: KER extraction based on a single compound. This mode starts with a specific target compound and aims to construct the overall toxicity mechanistic network associated with it. Its core task is to identify relationship chains containing at least three nodes from the literature: Target compound → Biological event A → Biological event B. The relationship between the target compound and the initial biological event is required to have an experimental basis or strong evidential support. For the relationships among subsequent biological events, even if the original text only proposes them as mechanistic hypotheses, they are allowed to be retained as long as they have reasonable biological significance, thereby enhancing the ability to discover potential mechanistic relationships. In addition, researchers can refine the literature topic by adjusting the retrieval query during the literature retrieval stage, and further specify specific toxicity endpoints or mechanistic directions in subsequent analyses using custom task prompts.

**Mode 2**: KER extraction based on a predefined event set. This mode is not restricted to a single compound, but centers on a predefined set of biological events defined by researchers for literature analysis. The predefined events can originate from existing AOPs, specific diseases, or research hypotheses. The system prioritizes identifying relationships among the predefined events while retaining other upstream and downstream events connected to the target events to discover under-explored mechanistic connections surrounding the research topic. This mode is suitable for mechanistic expansion of existing AOPs, evidence supplementation for specific KERs, and constructing topic-oriented networks around specific diseases or toxicity mechanisms.

Both modes support custom task prompts. Researchers can provide information such as toxicity endpoints, mechanistic directions, or targeted biological events based on specific research objectives. These prompts will be utilized in subsequent LLM analysis processes to ensure that literature screening and event extraction are conducted around the predefined research objectives.

## 2.2 Automated Literature Analysis and Data Output

After literature data are imported, the system first uses Python to parse the Excel file, extract the Title, Source, and Abstract fields, and read the corresponding DOI information of each publication. These are saved together with the literature records in a JSON data structure for subsequent processing. Subsequently, the system iteratively invokes the LLM to perform initial screening on the literature. Each publication is processed as an independent network request, and multithreading concurrency is used to improve analysis efficiency. The default number of threads is 20, which users can adjust based on the concurrency limits of the LLM API in use. The extracted DOI information is retained for subsequent KER result organization to link the original literature supporting the corresponding event relationships.

Literature screening is primarily based on the title and abstract to determine whether the publication meets the current research task. For Mode 1, the LLM determines whether the literature contains a relationship chain that satisfies the requirement of Target compound → Event A → Event B, focusing on evaluating whether the relationship between the target compound and the initial event has experimental or strong evidential support. For Mode 2, the LLM determines whether the literature involves target events or connections among related events based on the predefined event set and custom task prompts. This stage is completed solely by the LLM without invoking the knowledge retrieval function of AOP-Smart. Publications that pass the initial screening proceed to the subsequent KER extraction stage.

For publications that pass the screening, the system iteratively invokes the LLM to identify biological events and their upstream and downstream relationships. Mode 1 focuses on extracting the initial biological changes and their downstream developmental processes following exposure to the target compound; Mode 2 identifies relevant event relationships centered on the predefined event set while retaining other upstream and downstream events connected to the target events. The extraction results include upstream events (Source), downstream events (Target), as well as relevant species and literature information.

During this process, AOP-Smart acts as a connecting layer between the LLM and the official AOP-Wiki knowledge base, providing knowledge support for event identification and KE mapping. AOP-Smart constructs a local knowledge base based on official AOP-Wiki XML data and provides the LLM with official KE, KER, and AOP information through a Retrieval-Augmented Generation (RAG) mechanism.[8] After the LLM identifies biological events from the literature, the system calls AOP-Smart to retrieve relevant official KE information and assists in mapping based on the biological meaning of the events. For events that can be matched to official KEs, the corresponding official KE names and IDs are prioritized. For events that cannot be directly matched, the system further evaluates whether there are semantically similar official KEs or whether they are suitable for mapping to broader official event categories. When no reasonable official match exists, the original event description is retained and marked as a Candidate Key Event (Candidate KE). It should be noted that a Candidate KE indicates that no suitable official KE has been matched currently, which does not imply that the event has been confirmed as a new official KE.

After completing KER extraction for all publications, the system aggregates all Candidate KE names from the extraction results and submits them at once to the LLM for semantic standardization. Based on the biological meaning of each candidate event, the LLM identifies different terms used in various publications to describe the same or similar biological events, merges them into unified standardized names, and outputs a mapping dictionary between candidate events and standardized names according to a predefined format. Subsequently, Python uniformly replaces the Candidate KE names in the original KER data based on this dictionary and generates the final standardized results. This process only handles candidate events that have not yet been mapped to official KEs and does not modify the names of already matched official KEs.

After completing the above analyses, the system uses Python to aggregate all KER data and generate an Excel file. The output includes literature processing statistics, KE count statistics, KER count statistics, and a core relationship table for network construction. The core relationship table contains fields such as upstream events (Source), downstream events (Target), the number of supporting publications for the relationship (Count), relevant species (Species), and literature DOIs. Among these, Count represents the number of independent publications supporting the corresponding event relationship. For events mapped to official KEs, the system automatically adds URLs pointing to the official AOP-Wiki KE pages via Python. The DOI field is also set as a clickable link to facilitate researchers in tracing the original literature. The output relationship table is compatible with Cytoscape network analysis software and can be used for subsequent network construction, node filtering, and edge filtering. Furthermore, AOP-Smart employs a local knowledge base architecture based on official AOP-Wiki XML data, possessing good updatability and scalability. When AOP-Wiki releases new official XML data, users only need to replace the local XML file and re-execute the parsing to update the KE, KER, and AOP information in the local knowledge base, without modifying the core analysis pipeline of the system or manually maintaining and updating knowledge entries one by one. This design enables AOP-Smart to synchronize with official knowledge updates from AOP-Wiki in a timely manner, continuously providing the LLM with the latest standardized AOP domain knowledge, while providing a sustainably updatable knowledge foundation for subsequent KE mapping and mechanistic analysis in AOP-Smart-Finder.

### 2.3 Manual Curation, Network Analysis, and WoE Evaluation

After automated analysis is completed, researchers import the exported Excel relationship table into Cytoscape for network construction and manual curation. This primarily includes verifying official KE names and their corresponding IDs, checking the original literature via DOIs, identifying and removing unreasonable or redundant event relationships, and performing node merging, deletion, or supplementation based on specific research objectives. Additionally, researchers adjust the network structure according to research goals and perform toxicity classification of endpoint nodes. AOP-Smart-Finder does not automatically perform the toxicity classification of endpoint nodes, nor does it automatically determine the Cytoscape network layout.

After network verification is completed, researchers conduct a manual Weight of Evidence (WoE) evaluation of key KERs. This study evaluates them from three dimensions: Biological Plausibility, Empirical Support, and Essentiality.

**(1) Biological Plausibility**

Evaluates whether a clear and reasonable biological mechanistic link exists between the upstream and downstream events. High indicates support from clear and well-established biological mechanisms; Moderate indicates biological plausibility but the specific mechanism has not been fully elucidated; Low indicates only an event association without a clear mechanistic explanation.

**(2) Empirical Support**

Evaluates whether direct experimental support exists and whether the event relationships are consistent across different studies. High indicates consistent support under multiple experimental conditions, with temporal and dose concordance; Moderate indicates some experimental support, but the number of studies or evidence consistency is limited; Low indicates a lack of direct experimental support or the presence of clearly contradictory experimental results.

**(3) Essentiality**

Evaluates whether intervening in the upstream event can prevent or mitigate the occurrence of the downstream event. High indicates that blocking the upstream event can prevent or mitigate the downstream event; Moderate indicates indirect evidence that intervening in the upstream event affects the downstream event; Low indicates a lack of intervention evidence, or experimental results do not support the essentiality of the relationship.

Finally, by integrating the evaluation results of the above three dimensions, the overall WoE level for each KER is determined and classified as High, Moderate, or Low. All WoE evaluations are completed by researchers based on the original literature and relevant evidence, with evaluation principles referencing existing methodological studies on AOP WoE assessment.[9, 10]

## 3 Results

### 3.1 Case Study 1: High-Throughput Panoramic Analysis of DEHP-Driven Toxicity Mechanisms

To evaluate the processing capability of AOP-Smart-Finder on large-scale literature collections, Di(2-ethylhexyl) phthalate (DEHP) was selected as the target compound, and Mode 1 was employed to conduct an overall toxicity mechanism analysis without restricting toxicity endpoints. The Web of Science search query was: TS = ("DEHP" OR "Di(2-ethylhexyl) phthalate" OR "Bis(2-ethylhexyl) phthalate" OR 117-81-7). An initial search retrieved 12,983 literature records. The search results were imported into AOP-Smart-Finder, the single-compound KER extraction mode was selected, and no additional toxicity mechanism prompts were set to cover as many different toxicity mechanisms related to DEHP as possible. The analysis was conducted in October 2026 using the DeepSeek-Flash model with the thread count set to 200, and the AOP-Wiki official knowledge base version used was the 2026-10-01 XML release. The entire task was completed within 30 min, consuming a total of 53,730,793 input tokens and 19,799,749 output tokens.

Of the 12,983 total publications, the system skipped 11,890 and finally included 1,093 for analysis, extracting a total of 1,589 biological events and 2,519 KERs. Among them, 458 KEs were successfully mapped to the official AOP-Wiki knowledge base, while the remaining 1,131 were marked as Candidate KEs. To preliminarily evaluate the reliability of official KE mapping, 100 mapped results were randomly selected for manual curation. All were confirmed to be accurate, yielding a sampling verification accuracy of 100%. For events that could not be matched to suitable official KEs, the system retained the original descriptions to prevent the loss of potential biological information due to insufficient coverage of the current knowledge base.

Further analysis of the extracted network identified a total of 546 endpoint nodes with an out-degree of 0, involving reproductive toxicity (132), metabolic toxicity (123), endocrine toxicity (71), neurotoxicity (49), developmental toxicity (35), hepatotoxicity (30), immunotoxicity (30), cardiovascular toxicity (27), nephrotoxicity (14), respiratory toxicity (7), and other categories (28). Among these, endpoints related to reproductive and metabolic toxicity accounted for a relatively high proportion. The ultimately constructed overall toxicity mechanism network for DEHP is shown in Figure 2, illustrating the multi-level associations from compound exposure to different biological events and downstream toxicity endpoints, covering multiple tissues, organs, and toxic effects.

Because a large number of KERs (2,519) were obtained in this analysis, manual WoE evaluation was not performed on every single relationship. This case study primarily aims to demonstrate the application capability of AOP-Smart-Finder in large-scale literature processing, cross-literature biological event relationship discovery, and overall toxicity mechanism network construction, rather than to make a final judgment on the evidence strength of all extracted relationships.

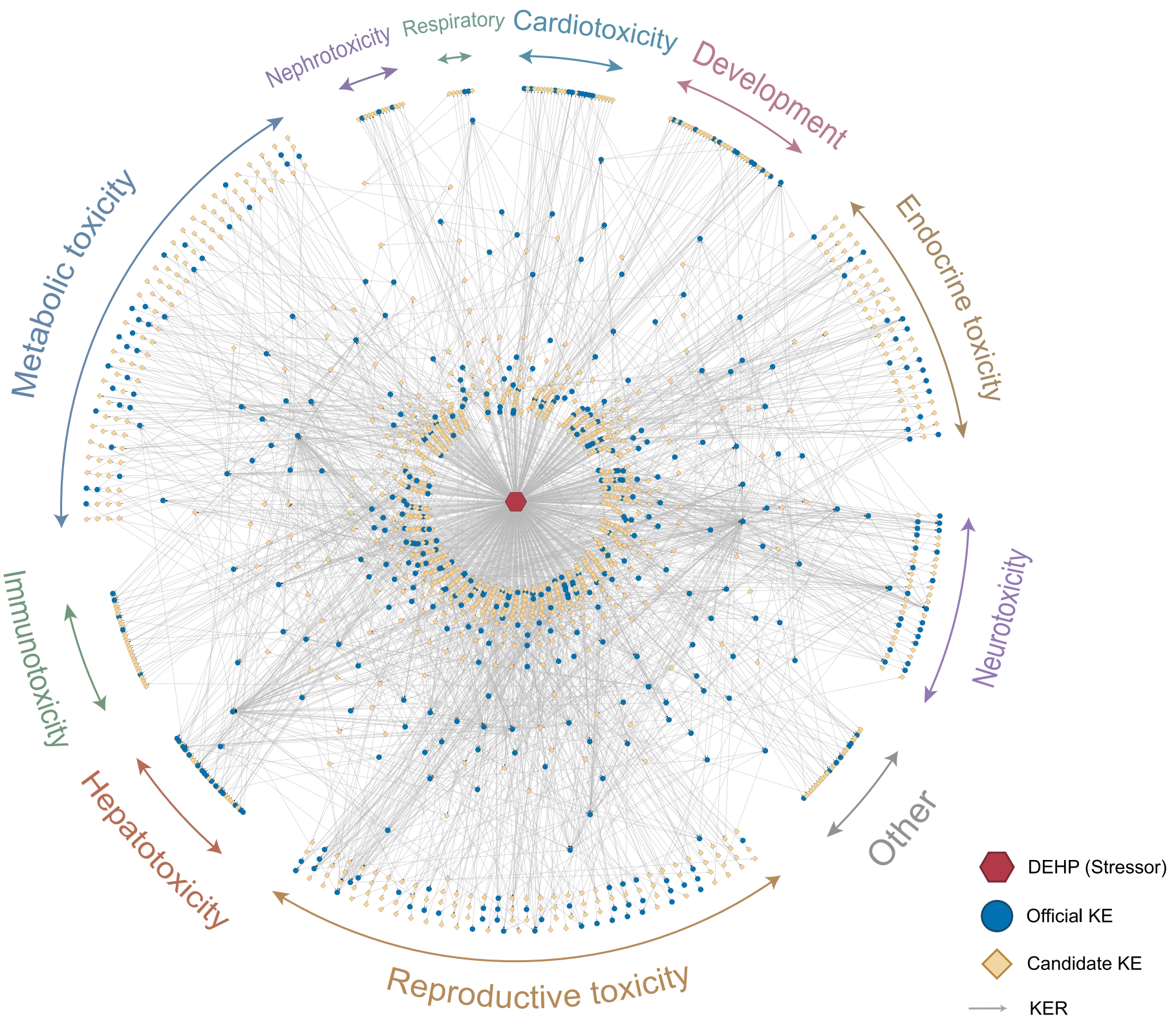

**Figure 2.** Overall toxicity mechanistic network of DEHP generated by AOP-Smart-Finder

### 3.2 Case Study 2: AOP-Informed Construction of the ALS Mechanistic Network

To further evaluate the application capability of AOP-Smart-Finder in mechanistic expansion and evidence supplementation for specific AOPs, the AOP 636 draft in AOP-Wiki was selected as the research reference. The title of this AOP is "Increase in reactive oxygen species (ROS) leading to human amyotrophic lateral sclerosis (ALS)". This AOP uses an increase in reactive oxygen species (ROS) as the initiating event and describes its connections with events such as oxidative stress, TDP-43 aggregation, mitochondrial dysfunction, and neuroinflammation, ultimately leading to human amyotrophic lateral sclerosis (ALS). However, the official AOP-Wiki page has not yet listed specific references for this AOP and lacks complete Overall Assessment information. Therefore, based on its existing mechanistic framework, this case study further identifies relevant event connections, supplements literature evidence for KERs, and conducts independent evidence evaluation through large-scale literature mining. It should be noted that this case study is not intended to re-establish or replace the official AOP 636, but rather to expand and supplement evidence for its existing mechanistic knowledge.

This case study employed Mode 2 to perform KER extraction without restricting compounds. Task prompts were set based on the core mechanisms of AOP 636, focusing on the potential connections among increased ROS, oxidative stress, TDP-43 aggregation, mitochondrial dysfunction, neuroinflammation, decreased neural network function, and ALS. The Web of Science search query was: TS = ("amyotrophic lateral sclerosis" OR "ALS") AND TS = ("oxidative stress" OR "reactive oxygen species" OR ROS).

An initial search retrieved 4,790 publications, and the top 500 ranked by relevance were selected as the analysis subjects. Among them, 437 publications completed automated literature analysis, while 63 were skipped by the system.

Analysis by AOP-Smart-Finder identified a total of 55 distinct stressors, including Paraquat, Rotenone, BMAA, Sodium arsenite, Malonate, Glyphosate, Aluminum, Copper, Iron, Cigarette smoke, Ionising radiation, and $H_2O_2$, encompassing chemicals, metals, and other environmental and experimental stimuli. In addition, the system extracted a total of 278 biological events, of which 85 were successfully mapped to official KEs in AOP-Wiki, and 548 event relationships were obtained.

To construct a more interpretable mechanistic network, the automated analysis results were imported into Cytoscape for manual curation. First, external stressor nodes and their associated connections were removed. Subsequently, relationships supported by at least 2 independent publications were retained, and isolated nodes generated after screening were deleted. On this basis, redundant or unreasonable event relationships were further manually reviewed and removed. The original Excel data were retained to allow for the subsequent recall of screened-out relationships based on research needs. Relationships supported by only 1 publication were not reintroduced in this case study, as the current screening results already formed a relatively complete mechanistic network.

Since each event relationship retained the corresponding number of supporting publications and DOIs, it was possible to directly trace the original studies and verify the biological meaning and evidential basis of the relationships one by one. Ultimately, a mechanistic network comprising 16 biological event nodes and 23 KERs was obtained, including 11 official KEs and 5 candidate events, with the relevant relationships supported by a total of 108 distinct publications.

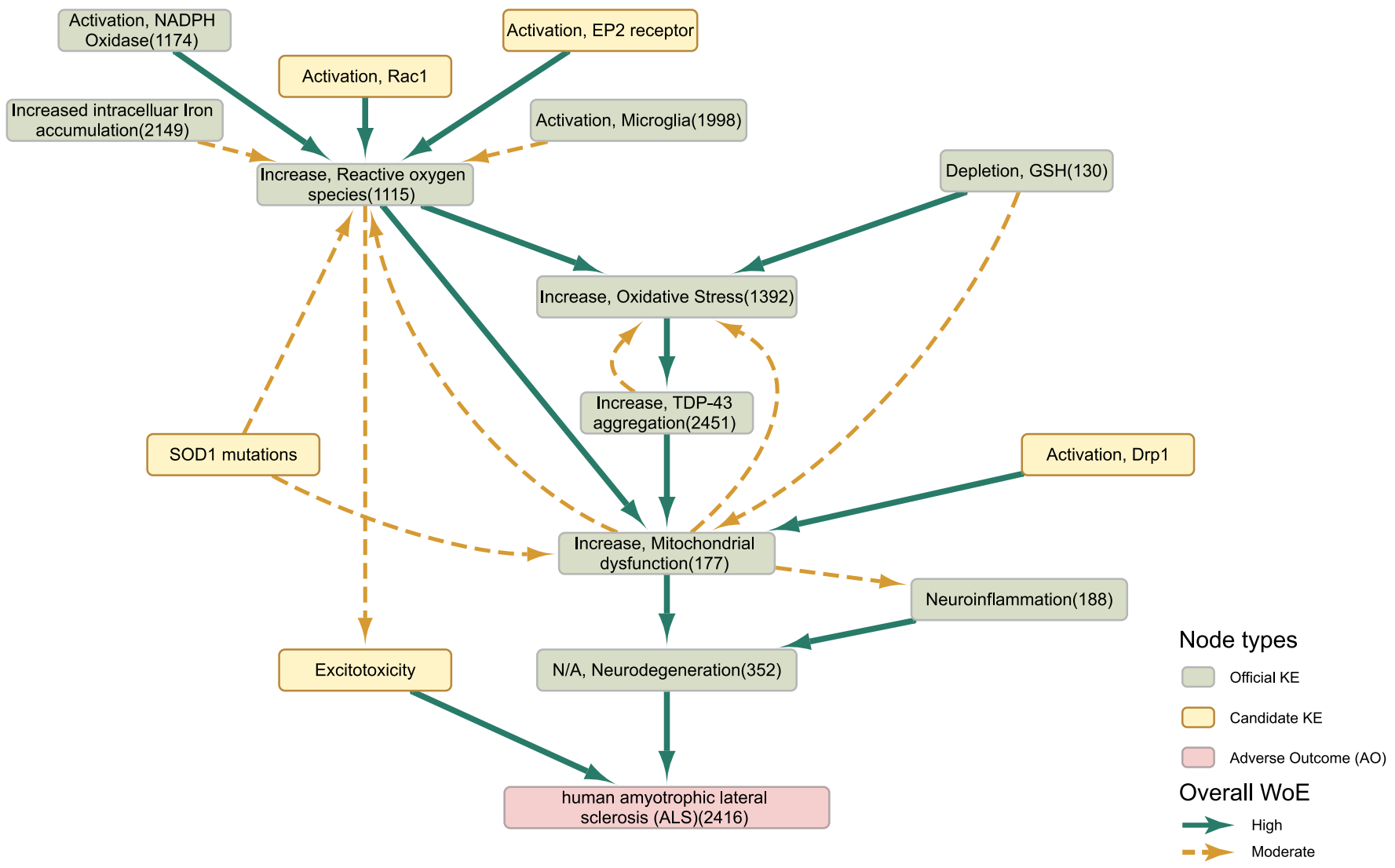

**Figure 3**. AOP 636-informed ALS mechanistic network

From the perspective of network structure, the final results not only retained the core initiating event of increased ROS in AOP 636 but also further identified multiple potential upstream regulatory factors and downstream mechanistic processes. At the upstream regulatory level, NADPH oxidase activation, Rac1 activation, EP2 receptor activation, and microglia activation all showed literature-supported connections with increased ROS. Among them, the NADPH oxidase/Rac1 signaling axis is associated with ROS generation and glial neurotoxicity induced by ALS-related SOD1 mutations.[11-14] EP2 receptor activation has also been confirmed to induce ROS production in motor neuron-like cells.[15] In addition, SOD1 mutations and intracellular iron accumulation were identified as upstream events associated with increased ROS.[16, 17] Among mitochondrial-related mechanisms, strong experimental support exists between Drp1 activation and mitochondrial dysfunction. Relevant studies have shown that inhibiting Drp1-mediated excessive mitochondrial fission can improve mitochondrial structure and function in ALS models and reduce ROS levels.[18, 19] Glutathione (GSH) depletion is associated with oxidative stress and mitochondrial dysfunction; existing studies have demonstrated that decreased GSH levels can exacerbate motor neuron injury.[20, 21]

In the ultimately constructed ALS mechanistic network, a biologically significant bidirectional association structure was formed between increased ROS and mitochondrial dysfunction. This structure reflects a potentially mutually reinforcing and continuously amplifying process between oxidative damage and mitochondrial abnormalities. On the one hand, excessive ROS production can affect mitochondrial respiration and energy metabolism by oxidatively damaging mitochondrial components. Beretta et al.[22] found in cell models expressing ALS-related G93A-SOD1 mutant protein that both cytosolic and mitochondrial ROS levels were significantly elevated, accompanied by decreased ATP levels and mitochondrial dysfunction; further use of the antioxidant N-acetylcysteine (NAC) to reduce ROS restored mitochondrial functional indicators to control levels, providing experimental support for ROS-mediated mitochondrial damage. On the other hand, mitochondrial dysfunction itself may also become an important source of sustained ROS production. Abnormal function of the mitochondrial electron transport chain can alter electron transfer efficiency, increasing electron leakage and ROS generation. Panov et al. [23] found in G93A-SOD1 transgenic rats that the levels of spinal cord mitochondrial respiratory chain complexes III and IV decreased by approximately 60% and 35%, respectively, while mitochondrial ROS production was significantly increased, reaching 5–7 times that of normal controls. The authors further noted that abnormal ROS generation may promote motor neuron degeneration by accelerating mitochondrial damage. The above studies support the interaction between ROS and mitochondrial dysfunction from different directions. Furthermore, Szeto et al.[24] also pointed out that ROS generation and mitochondrial functional damage can form a continuously amplifying, self-sustaining feedback process. Therefore, this study integrated the bidirectional relationship between increased ROS and mitochondrial dysfunction into a potential positive feedback loop, suggesting that the two may form a sustained pathological amplification mechanism during ALS-related oxidative damage.

From the WoE evaluation results, the strength of evidence for these two directions is not entirely identical. The relationship from increased ROS to mitochondrial dysfunction was rated High across all three dimensions of biological plausibility, empirical support, and essentiality, with an overall WoE of High. In contrast, although the relationship from mitochondrial dysfunction to increased ROS has high biological plausibility, its empirical support and essentiality were both Moderate, resulting in an overall WoE of Moderate. This indicates that the evidence for ROS-mediated mitochondrial functional damage is relatively sufficient, whereas the feedback effect of mitochondrial abnormalities further promoting ROS generation still requires more direct intervention experiments for verification. The two relationships together constitute the core feedback module in the network, also demonstrating that combining cross-literature event relationships and multi-dimensional WoE evaluation can further reveal complex biological connections that are difficult to fully express through unidirectional mechanistic chains.

Notably, the system also identified a bidirectional association between oxidative stress and TDP-43 aggregation, as well as a connection between TDP-43 aggregation and mitochondrial dysfunction. The relationship in which oxidative stress promotes TDP-43 aggregation had an overall WoE of High, while the reverse relationship where TDP-43 aggregation affects oxidative stress was Moderate. The relationship between TDP-43 aggregation and mitochondrial dysfunction received relatively high evidential support. Zuo et al.[25] found that oxidative stress can promote TDP-43 aggregation, and abnormally aggregated TDP-43 in turn causes mitochondrial homeostasis imbalance by affecting the expression and function of mitochondria-related proteins, further exacerbating oxidative stress. This provides direct experimental evidence for the interconnections among oxidative stress, TDP-43 aggregation, and mitochondrial abnormalities revealed by this network. Other studies also support the connections among these events from aspects such as glutathione intervention, antioxidant protection, and TDP-43-related pathological changes[26-29].

At the downstream mechanistic level, mitochondrial dysfunction further connects to neuroinflammation and neurodegeneration, and strong literature support exists between neuroinflammation and neurodegeneration. ALS-related studies have shown that the inflammatory response of astrocytes can be accompanied by abnormalities in ROS, inflammatory cytokines, and glutamate regulation, causing damage to motor neurons.[30] In addition, the system identified connections between increased ROS and excitotoxicity, as well as downstream connections among excitotoxicity, neurodegeneration, and the ALS outcome. Both excitotoxicity and oxidative stress are considered to be involved in the pathological process of ALS.[31, 32] Consequently, starting from the initial ROS disturbance, the final network further connects multiple mechanistic components, including mitochondrial abnormalities, oxidative stress, TDP-43 protein pathology, neuroinflammation, and neurodegeneration, forming a branched network leading to the ALS outcome.

After completing the manual curation of the network, WoE evaluation was performed on the 23 KERs ultimately retained. The biological plausibility of all relationships was rated as High; in the overall evaluation, 13 KERs were rated as High, 10 as Moderate, and no relationship was rated as Low. Among them, the relationship between increased ROS and mitochondrial dysfunction was rated High across all three dimensions of Biological Plausibility, Empirical Support, and Essentiality, supported by 12 publications. The reverse relationship between mitochondrial dysfunction and increased ROS was also supported by 12 publications; however, because its empirical support and essentiality were both Moderate, its final overall evaluation was also Moderate. This result indicates that the number of supporting publications cannot solely determine the overall evidential strength of a KER; it requires comprehensive judgment combined with the consistency of experimental results, mechanistic explanations, and intervention evidence.

Overall, based on the existing core mechanisms of AOP 636, this case study further identified and integrated multiple upstream regulatory events, oxidative stress-related processes, and downstream neural injury mechanisms, forming a literature-supported mechanistic network composed of 16 event nodes and 23 relationships. By combining large-scale literature mining, official KE mapping, manual network curation, and WoE evaluation, AOP-Smart-Finder can not only assist in expanding the mechanistic structure of existing AOPs but also further identify the evidential strengths and weaknesses of different mechanistic relationships, providing a basis for subsequent mechanistic research and AOP refinement.

**Table 1**. Manual WoE evaluation results of the 23 KERs in the ALS mechanistic network.

| Upstream Event → Downstream Event | Weight of evidence | | | |
|---|---|---|---|---|
| | Biological Plausibility | Empirical Support | Essentiality | Overall WoE |
| Activation, NADPH Oxidase(1174)→Increase, Reactive oxygen species(1115) | High | High | High | **High** |
| Increased intracelluar Iron accumulation(2149)→Increase, Reactive oxygen species(1115) | High | Moderate | Low | **Moderate** |
| Activation, Rac1→Increase, Reactive oxygen species(1115) | High | High | High | **High** |
| Activation, EP2 receptor→Increase, Reactive oxygen species(1115) | High | High | High | **High** |
| Activation, Microglia(1998)→Increase, Reactive oxygen species(1115) | High | High | Moderate | **Moderate** |
| SOD1 mutations→Increase, Reactive oxygen species(1115) | High | High | Moderate | **Moderate** |
| SOD1 mutations→Increase, Mitochondrial dysfunction(177) | High | Moderate | Low | **Moderate** |
| Increase, Reactive oxygen species(1115) → Increase, Mitochondrial dysfunction(177) | High | High | High | **High** |
| Increase, Mitochondrial dysfunction(177) → Increase, Reactive oxygen species(1115) | High | Moderate | Moderate | **Moderate** |
| Activation, Drp1→Increase, Mitochondrial dysfunction(177) | High | High | High | **High** |
| Increase, Reactive oxygen species(1115)→Increase, Oxidative Stress(1392) | High | High | Moderate | **High** |
| Increase, Mitochondrial dysfunction(177)→Increase, Oxidative Stress(1392) | High | Moderate | Moderate | **Moderate** |
| Depletion, GSH(130)→Increase, Oxidative Stress(1392) | High | High | Moderate | **High** |
| Depletion, GSH(130)→Increase, Mitochondrial dysfunction(177) | High | High | Moderate | **Moderate** |
| Increase, Mitochondrial dysfunction(177)→Neuroinflammation(188) | High | Moderate | Low | **Moderate** |
| Increase, Mitochondrial dysfunction(177)→N/A, Neurodegeneration(352) | High | High | Moderate | **High** |
| Increase, Oxidative Stress(1392)→Increase, TDP-43 aggregation(2451) | High | High | Moderate | **High** |
| Increase, TDP-43 aggregation(2451)→Increase, Oxidative Stress(1392) | High | Moderate | Moderate | **Moderate** |
| Increase, TDP-43 aggregation(2451) → Increase, Mitochondrial dysfunction(177) | High | High | Moderate | **High** |
| Increase, Reactive oxygen species(1115)→Excitotoxicity | High | Moderate | Low | **Moderate** |
| Excitotoxicity→human amyotrophic lateral sclerosis (ALS)(2416) | High | High | Moderate | **High** |
| N/A, Neurodegeneration(352) → human amyotrophic lateral sclerosis (ALS)(2416) | High | High | Moderate | **High** |
| Neuroinflammation(188)→N/A, Neurodegeneration(352) | High | High | Moderate | **High** |

**Note:** Numbers in parentheses indicate the official KE IDs. Interestingly, the spelling "intracelluar" (for KE 2149) is retained exactly as it appears in the official AOP-Wiki database. This is because the local knowledge base used by AOP-Smart was built on the 2026-10-01 XML release, and we also verified

on the official AOP-Wiki webpage (accessed on November 5, 2026) that this spelling remains unchanged. This demonstrates that when the LLM integrates with AOP-Smart, it strictly adheres to the principle of directly utilizing official IDs and terminology from the local knowledge base, without altering the original text.

## 4 Discussion

### 4.1 High-Throughput Literature Analysis Provides a New Avenue for AOP Mechanism Discovery

The establishment and refinement of AOPs rely on the systematic integration of biological events and their relationships from a large number of dispersed publications. Traditional manual screening and evidence organization methods are time-consuming and struggle to meet the demands of large-scale literature analysis. AOP-Smart-Finder achieves efficient analysis of large-scale literature by processing literature screening and KER extraction in separate stages and incorporating multithreading concurrency. In the DEHP case, the system processed 12,983 literature records within 30 min and extracted 1,589 biological events and 2,519 KERs from the 1,093 publications ultimately included, providing a high-throughput analytical pathway for exploring the overall toxicity mechanisms of a single compound.

Notably, only 458 of the 1,589 events could be mapped to official KEs in AOP-Wiki, accounting for 28.82%, while the remaining 1,131 (71.18%) were retained as Candidate KEs. This result reflects a certain degree of insufficient coverage by the existing official KE knowledge system regarding the biological events involved in large-scale literature. Although candidate events may include terminological variations or events that have not yet been fully validated and cannot be directly regarded as new official KEs, they provide important clues for discovering potential knowledge gaps and indicate that existing AOP knowledge still has room for further expansion and refinement.

AOP-Smart-Finder can rapidly identify candidate biological events and their upstream/downstream relationships from the literature, complete official KE mapping via AOP-Smart, and simultaneously retain unmatched events along with their original literature evidence, providing a basis for subsequent manual curation, event standardization, and AOP-Wiki knowledge supplementation. In addition, a total of 546 endpoint nodes with an out-degree of 0 were identified in the DEHP network, covering multiple toxicity domains. Overall, this case study demonstrates that AOP-Smart-Finder can not only improve the efficiency of large-scale literature analysis but also help identify potential coverage gaps in existing AOP knowledge, providing support for the continuous expansion and updating of AOP knowledge.

### 4.2 Dual-Mode Design Expands the Application Scope of AOP-Smart-Finder

Another key feature of AOP-Smart-Finder is that it provides two distinct analysis modes. Mode 1 centers on a single compound, emphasizing the extraction of relationship chains from exogenous chemicals to biological events and subsequently to downstream toxicity mechanisms. This mode is suitable for investigating the overall toxicity mechanisms of compounds and is capable of exploring various biological processes that may be involved without predefining the toxicity direction. Mode 2, in contrast, does not rely on a single compound but centers on an event set provided by the user, using existing AOPs, disease-related mechanisms, or specific biological processes as the analytical basis.

The ALS case study demonstrates the practical value of this mode. By focusing on ROS and ALS in AOP 636, the system not only extracted relationships directly related to the core events but also identified other relevant events, such as SOD1 mutations, Drp1 activation, TDP-43 aggregation, and excitotoxicity. These events and their relationships enriched the original mechanistic network and provided traceable literature sources for further investigation.

It should be emphasized that the final network in the ALS case is not a simple copy of the original AOP 636 structure, but an AOP-informed mechanistic network formed through large-scale literature integration, using existing knowledge as a starting point. This approach is highly scalable and can be further applied to the mechanistic enrichment of other registered AOPs, KER evidence supplementation, and specific disease mechanism studies.

### 4.3 AOP-Smart Bridges LLM Extractions with Official AOP Knowledge

In AOP-Smart-Finder, the role of AOP-Smart is not only to provide an external reference database, but more importantly, to establish a connecting mechanism between LLM outputs and official AOP knowledge[8]. If LLMs were solely used to directly identify KEs, the model might express the same biological event in different ways or generate non-existent official KE IDs, making it difficult for subsequent networks to align with the existing AOP-Wiki knowledge system. AOP-Smart constructs a searchable knowledge environment based on local AOP-Wiki XML data, enabling the model to reference official KE information during the event extraction stage and thereby prioritize the use of existing standard KE names and IDs. Meanwhile, the design of retaining Candidate KEs avoids excessive reliance on the official knowledge base. If a biological event cannot be mapped to an existing official KE, the system will not simply delete it but will retain its original description, allowing it to enter subsequent standardization and manual curation processes as a candidate event. This approach strikes a balance between official knowledge consistency and the discovery of new events. Furthermore, as the AOP-Wiki knowledge base is continuously updated, researchers only need to replace the local XML file to introduce the latest official KE standards into subsequent analyses. This design helps maintain consistency in event mapping across different batches of literature analysis.

### 4.4 Manual Curation and WoE Evaluation Are Indispensable Components of the Framework

AOP-Smart-Finder is not a fully autonomous AOP generation system, but rather a human-AI hybrid mechanistic discovery framework. Although LLMs can rapidly identify biological events and their relationships in the literature, direct experimental results, mechanistic hypotheses, and statistical associations in publications possess varying strengths of evidence, and automatically extracted event relationships still require further verification. Therefore, this study places manual review after automated analysis, utilizing the DOIs and official KE information retained in the results to trace the original literature and evaluate the events and their upstream/downstream relationships. In the ALS case, researchers first performed a preliminary screening of the extraction results based on the number of supporting publications and then used Cytoscape to further remove redundant or unreasonable event relationships and isolated nodes, obtaining a relatively streamlined mechanistic network. On this basis, manual WoE evaluation was conducted on the ultimately retained KERs from three dimensions: biological plausibility, empirical support, and essentiality, to differentiate the evidence strength of various mechanistic relationships. Therefore, the primary value of AOP-Smart-Finder lies in using automated analysis to improve the efficiency of discovering and integrating potential mechanistic relationships, while ensuring the scientific validity of the final mechanistic network through manual curation and WoE evaluation. This division of labor effectively combines high-throughput literature mining with scientific judgment in specialized domains.

### 4.5 Limitations and Future Perspectives

Although AOP-Smart-Finder demonstrated good application capabilities in the two case studies, there is still room for further refinement. Currently, literature analysis is primarily based on the titles and abstracts exported from Web of Science and has not yet fully utilized the experimental results, figures, and supplementary material information from the full texts. In the future, full-text analysis and key evidence paragraph extraction can be further introduced to improve the completeness of identifying biological events and their relationships.

Overall, AOP-Smart-Finder integrates large-scale literature screening, biological event extraction, official knowledge mapping, and manual evidence evaluation into a comprehensive analytical framework, providing a new technical pathway for cross-literature mechanism discovery and AOP knowledge expansion. With the further refinement of full-text analysis capabilities and quantitative evaluation systems, its application potential in complex biological mechanism research and AOP knowledge updating is expected to be further realized.

## 5 Conclusions

This study proposes AOP-Smart-Finder, a high-throughput human-AI hybrid framework for AOP discovery and evidence mapping from large-scale scientific literature. By integrating LLM-based literature screening and KER extraction, AOP-Smart official KE retrieval and mapping, Python data organization, as well as manual network curation and WoE evaluation, the framework achieves a continuous analytical workflow from literature collections to structured mechanistic networks. Through two analysis modes based on a single compound and a predefined event set, the system can accommodate various research needs, including overall toxicity mechanism exploration, existing AOP mechanistic expansion, and relationship analysis of specific biological events. The DEHP case demonstrated the framework's capability to process ultra-large-scale literature collections and construct overall toxicity mechanistic networks; the ALS case further showcased its application value in using existing AOP knowledge to guide literature mining, supplement mechanistic connections, and trace KER evidence. More importantly, AOP-Smart-Finder can not only utilize existing official KEs for knowledge mapping but also retain unmatched candidate events, providing informational support for the identification of potential new KEs and the continuous refinement of existing AOP knowledge. By combining automated literature mining with manual evidence review, AOP-Smart-Finder provides a scalable, traceable, and human-AI hybrid solution for large-scale literature-driven AOP discovery, mechanistic network construction, and knowledge integration.